\documentclass[10pt,twocolumn,letterpaper]{article}

\usepackage[pagenumbers]{cvpr}
\usepackage{times}

\input{pidinner/preamble}
\usepackage{placeins} 

\definecolor{pidrsilink}{rgb}{0.21,0.49,0.74}

\usepackage[
    pagebackref,
    breaklinks,
    colorlinks,
    allcolors=pidrsilink
]{hyperref}

\title{Improving the Improver: Semantic PID Control for \\Medical Agent Self-Improvement}

\author{%
  Dexia Chen\textsuperscript{1,*},
  Yilong Hu\textsuperscript{1,*},
  Wenqi Zhong\textsuperscript{1,2,*},
  Guangyu Guo\textsuperscript{1,2,*},\\
  Tony C. W. MOK\textsuperscript{1},
  Ling Zhang\textsuperscript{1},
  Zhongwei Qiu\textsuperscript{1,2,\textdagger}\\[0.35em]
  \textsuperscript{1}DAMO Academy, Alibaba Group
  \qquad
  \textsuperscript{2}Zhejiang University\\[0.25em]
  {\small\ttfamily
    \{chendexia.cdx, boyue.hyl, qiuzhongwei.qzw\}@alibaba-inc.com}\\[0.2em]
  {\small
    \textsuperscript{*}These authors contributed equally.
    \quad
    \textsuperscript{\textdagger}Corresponding author.}%
}
\begin{document}

\maketitle

\begin{abstract}
High-stakes vertical medical tasks, such as postoperative pathological staging and long-term survival prediction, demand multi-step clinical reasoning over heterogeneous evidence under weak supervision and high outcome uncertainty.
While we first orchestrate \textbf{CliAgent}, a domain-tailored clinical agent workflow that systematically grounds predictions in multi-modal patient evidence, manually designed medical agents remain constrained by static, brittle rules. Moreover, existing recursive self-improvement (RSI) methods suffer from \emph{feedback ambiguity} and severe \emph{development-set overfitting} when evolving agents in high-noise medical domains. Focusing on \textbf{improving the improver}, we propose \textbf{Semantic PID-RSI}, an evidence-guided control-theoretic framework for medical agent self-improvement. Inspired by PID control, our improver evaluates candidate modification batches via three diagnostic roles: \textbf{P} isolates current clinical failure modes, \textbf{I} synthesizes longitudinal evidence from past trials, and \textbf{D} estimates response dynamics and regression risks (\textbf{D-pre}/\textbf{D-post}). Selected plans are compiled into \textbf{behavior contracts} for staged probing and independent adoption judging. Instantiated and validated on challenging oncology tasks (pT/pN staging and survival prognosis), evolving \textbf{CliAgent} with \textbf{Semantic PID-RSI} on a \textbf{3B-activated} open-weight model achieves substantial gains (up to \textbf{+6.27} C-index points), outperforming overfitting-prone RSI baselines and approaching the performance of frontier proprietary models (e.g., GPT-6 series) equipped with \textbf{CliAgent}.
\end{abstract}
\section{Introduction}

High-stakes vertical medical tasks, exemplified by postoperative pathological staging (e.g., pT and pN staging) and long-term survival prediction in oncology~\cite{amin2017ajcc,smyth2020gastric}, carry profound implications for clinical treatment planning. Unlike routine medical question answering~\cite{singhal2023large,nori2023capabilities}, decision-making in specialized clinical domains demands multi-step reasoning over noisy, heterogeneous patient evidence under weak supervision and high outcome uncertainty~\cite{jiang2021development,chen2022pan}. When applied directly, general-purpose large language models (LLMs) struggle to systematically weigh conflicting clinical indicators~\cite{hager2024evaluation}. To structure clinical decision-making beyond raw prompting, building on recent advances in medical agentic workflows~\cite{tang2024medagents,li2024agent,tu2025towards}, we first orchestrate {CliAgent}, a domain-tailored clinical agent workflow that systematically grounds predictions in multi-modal patient evidence through structured extraction, rule verification, and risk aggregation, establishing a reliable reasoning foundation across diverse LLM backbones.

Despite providing a strong structural prior, manually orchestrated medical agents like {CliAgent} remain bottlenecked by static, human-crafted aggregation rules that are brittle across complex patient subpopulations. Recursive self-improvement (RSI) and automated agent optimization methods, ranging from verbal self-reflection~\cite{pidrsi_shinn2023reflexion,madaan2023self} and prompt evolution~\cite{yang2024large,khattab2024dspy,yuksekgonul2025optimizing} to executable workflow and program search~\cite{hu2025automated,zhang2025aflow,pidrsi_novikov2025alphaevolve,zhang2026agentic}, offer a promising paradigm to automatically refine domain prompts and agentic structures using evaluator feedback. However, when evolving medical agents in high-noise clinical environments, the effectiveness of RSI is fundamentally bottlenecked by the quality of the \emph{improver} itself, exposing two critical failure modes. First, \textbf{feedback ambiguity}: an aggregate scalar score (e.g., AUC or Harrell's C-index~\cite{harrell1982evaluating}) indicates whether performance shifted, but leaves ambiguous which specific clinical reasoning step caused the error and whether a similar edit already failed in earlier rounds. Second, \textbf{uncontrolled regression and development-set overfitting}: without explicit behavioral constraints, existing improvers greedily exploit spurious correlations on the development split while silently breaking valid clinical logic. For example, state-of-the-art RSI frameworks such as OPRO~\cite{yang2024large}, Penguin-RSI, and ACE~\cite{zhang2026agentic} achieve high development scores (up to 74.35\% Avg.) but suffer severe generalization drops of 4.86--5.39 points on the held-out test cohort in our oncology tasks.

To overcome these challenges, we shift focus to {improving the improver} and propose \textbf{Semantic PID-RSI}, an evidence-guided control-theoretic framework that recursively evolves {CliAgent}. Inspired by classic Proportional-Integral-Derivative (PID) control in dynamical systems~\cite{astrom2005advanced}, our improver replaces unconstrained reflection with three parallel, complementary diagnostic analyses over a shared batch of candidate modification plans. Specifically, {Proportional (P) current-error analysis} isolates concrete clinical failure modes observed in the active agent and grounds each proposal in a specific performance gap; {Integral (I) historical-evidence analysis} synthesizes longitudinal records from both successful and unsuccessful prior trials, preventing repetitive dead-ends while identifying when changed conditions justify revisiting an earlier hypothesis; and {Derivative (D) response-and-risk analysis} estimates intervention scope, dependencies, and regression risks prior to execution ({D-pre}), while reconciling expected versus realized behavioral shifts after evaluation ({D-post}). Evaluating a common candidate batch through these three lenses makes immediate error correction, long-term trial memory, and regression risks directly comparable before allocating evaluation budget.

Furthermore, {Semantic PID-RSI} enforces disciplined workflow evolution through {behavior contracts} and {independent adoption}. Rather than applying free-form edits to {CliAgent}, each selected plan is compiled into a formal behavior contract~\cite{meyer1992applying} specifying the intended behavioral delta, invariant behaviors to preserve, and explicit falsification criteria. Revised agents undergo staged probing followed by full development-set evaluation, after which an independent adoption judge evaluates the measured response against the contract to decide whether to update the incumbent program. Both adopted and rejected trials, together with their {D-post} discrepancy analyses, are archived into a structured research history, closing the loop between hypothesis formulation and empirical evidence.

Instantiated and validated on three challenging oncology tasks (namely cancer pT staging, pN staging, and survival prognosis), evolving CliAgent with Semantic PID-RSI on a lightweight 3B-activated open-weight model achieves substantial held-out test gains (up to +6.27 C-index points over the initial agent). Reaching the highest held-out average among the compared RSI frameworks (70.30\%), our 3B-activated medical agent effectively curbs development-set overfitting, outperforming competing RSI baselines and even approaching frontier proprietary models equipped with CliAgent.

Our contributions are summarized as follows:
\begin{itemize}
    \item We introduce CliAgent, a clinical agent workflow that combines report and CT evidence through structured extraction and aggregation for pathological staging and survival prediction.
    \item We propose Semantic PID-RSI, a control-inspired framework that assesses candidate modifications through current-error analysis (P), historical-evidence analysis (I), and response-and-risk analysis before and after evaluation (D-pre/D-post).
    \item We combine behavior contracts, staged evaluation, independent adoption judging, and experience reuse to connect proposed changes with their observed effects and preserve evidence from both adopted and rejected trials.
    \item On gastric-cancer prediction tasks, Semantic PID-RSI improves the held-out three-task average by 2.92 points and survival C-index by 6.27 points over initialization, achieving the highest held-out average among the compared RSI frameworks.
\end{itemize}
\section{Related Work}

\subsection{Agent and Workflow Optimization}
    Program and workflow search uses measured outcomes to guide structural or prompt-level changes to task-solving agents. OPRO treats language models as black-box optimizers guided by scalar scores~\cite{yang2024large}, while DSPy
     compiles declarative language pipelines into optimizable modules~\cite{khattab2024dspy} and TextGrad backpropagates textual feedback to refine system components~\cite{yuksekgonul2025optimizing}. Extending optimization to
     executable structures, FunSearch search over code modifications with automated evaluators~\cite{romera2024mathematical}, and ADAS search over agentic architectures 
    represented as code or computational graphs~\cite{hu2025automated}. GEPA instead evolves prompts through natural-language reflection on execution trajectories with Pareto-aware 
    selection~\cite{pidrsi_agrawal2026gepa}. 
    A parallel line studies recursive self-improvement~\cite{pidrsi_gao2026survey}, where STOP improves its own optimization scaffolding~\cite{zelikman2023self} and G\"odel Agent formalizes self-referential improvement~\cite{pidrsi_yin2025godelagent}. 
    These methods establish evaluator-guided optimization, but aggregate task scores or free-form textual gradients often leave \emph{feedback ambiguity}: the same outcome can motivate several different
     changes, inviting redundant exploration and development-set overfitting. \PIDRSIMethod{} addresses this gap by combining current-error (P), historical-evidence (I), and response-risk (D) analyses of a common candidate batch before any program edit is committed, while the search procedure and underlying models remain fixed.

    \subsection{Feedback and Long-Term Memory}
    Language-mediated feedback and memory let agents refine behavior across trials. Self-Refine alternates generation, self-feedback, and revision within a single task instance~\cite{pidrsi_madaan2023selfrefine}, and Reflexion persists 
    verbal feedback in an episodic buffer to inform subsequent trials~\cite{pidrsi_shinn2023reflexion}. Beyond trial-level reflection, Voyager accumulates an ever-growing library of executable skills~\cite{wang2023voyager}, and 
    ExpeL extracts cross-task natural-language insights from both successful and failed trajectories~\cite{zhao2024expel}. These mechanisms show that structured verbal experience can guide later decisions, yet the feedback is 
    typically attached to outcomes rather than to explicit intervention hypotheses. 
    In contrast, Semantic PID-RSI anchors feedback to explicit intervention hypotheses governed by behavior contracts. By archiving both adopted and rejected trials alongside pre- and post-evaluation discrepancy analyses, our framework prevents redundant exploration and distinguishes repeated failures from hypotheses tested under shifted conditions.

\subsection{Autonomous Research and Medical Agents}
    Recent work automates the scientific discovery lifecycle and deploys multi-agent systems in clinical domains. The AI Scientist and Agent Laboratory automate research workflows spanning idea generation, experimentation, and manuscript
     preparation~\cite{pidrsi_lu2024aiscientist,schmidgall2025agent}. In medicine, MedAgents organizes fixed multi-disciplinary role-playing for zero-shot reasoning~\cite{tang2024medagents}, and MDAgents adaptively 
    selects collaboration structures by task complexity, though its decision rules remain manually designed rather than learned from execution evidence~\cite{kim2024mdagents}. AgentClinic benchmarks agents in simulated clinical 
    environments~\cite{schmidgall2024agentclinic}, and Agent Hospital evolves doctor agents from simulated medical records via its MedAgent-Zero strategy~\cite{li2024agent}. These efforts target open-ended research 
    automation, clinical collaboration, or simulation-based evolution. \PIDRSIMethod{} instead concentrates on the recursive self-improvement cycle of a medical task program itself: formulating a modification hypothesis, revising the 
    executable workflow under behavior contracts, evaluating staged behavior on real gastric-cancer pT staging, pN staging, and survival prediction, and preserving the resulting longitudinal evidence. By holding model weights, 
    search-role definitions, and the independent adoption criterion fixed, it isolates and improves the \emph{improver}, enabling reliable evolution under high clinical uncertainty.

\section{Method}

We introduce \PIDRSIMethod{}, a framework for improving clinical agents through structured analysis of proposed program changes (Fig.~\ref{pidrsi:fig:overview}). We initialize the agent with CliAgent, which combines report and CT evidence for pathological staging and survival prediction. At each search round, Semantic PID assesses candidate plans using current errors, previous trials, and anticipated response and risk. Selected plans are implemented under behavior contracts and evaluated before an independent judge decides whether to update the current program. Both accepted and rejected trials inform subsequent search.

\subsection{Clinical Agent and Problem Formulation}

Given a patient case $x=(R,V)$ comprising a report $R$ and CT images $V$, CliAgent produces predictions for pathological T staging, pathological N staging, and survival prognosis. Its task program $h$ specifies the instructions, evidence use, and coordination of clinical reasoning. A fixed executor $E_{\theta}$ runs the program to produce $\hat{\mathbf{y}}=E_{\theta}(x;h)$.

\begin{figure*}[t]
\centering
\includegraphics[width=0.94\textwidth]{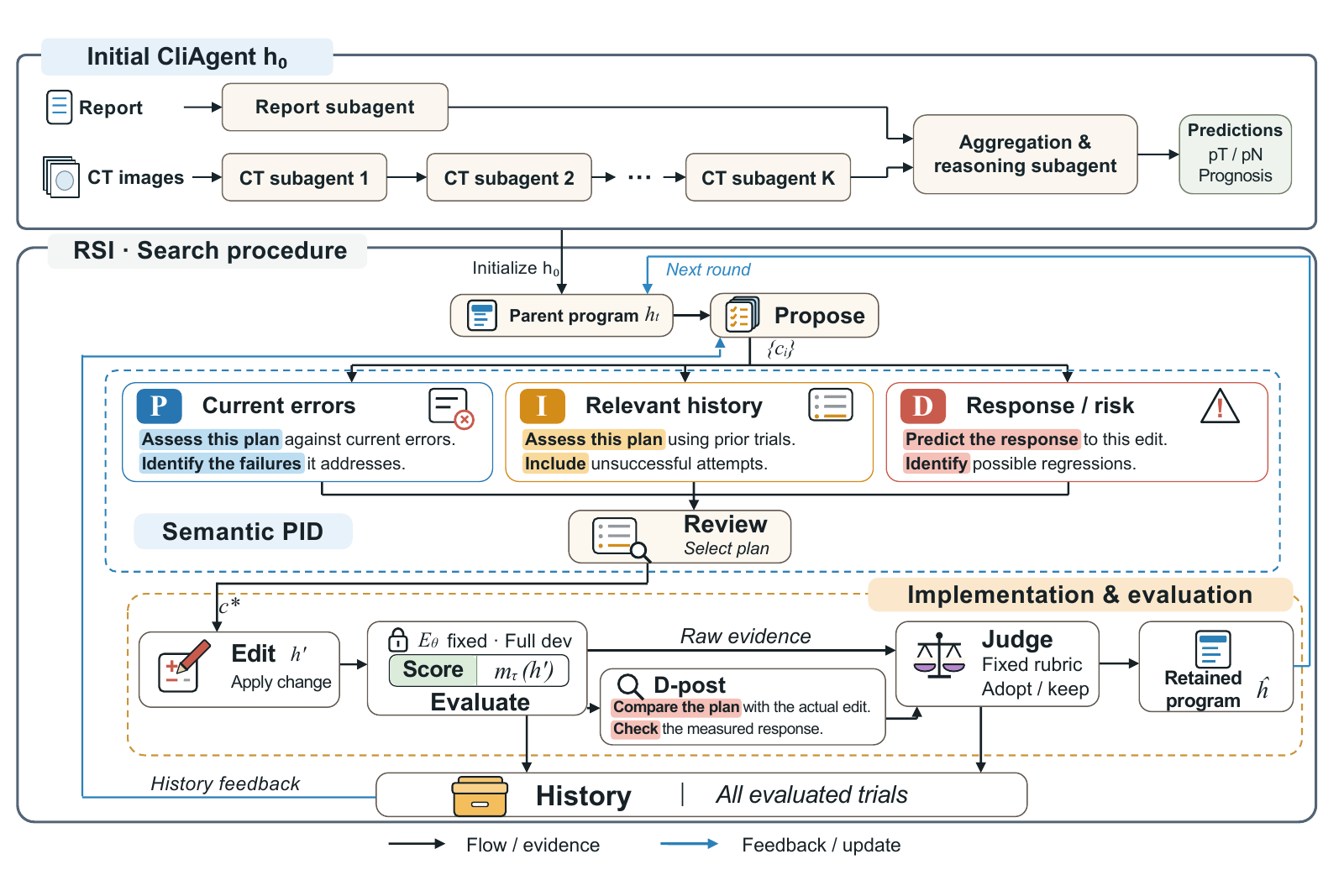}
\vspace{-0.8cm}
\caption{\textbf{Overview of Semantic PID-RSI.} Initialized with the multi-modal CliAgent workflow ($h_0$), each optimization round proposes candidate modification plans $\{c_i\}$ and evaluates them in parallel via Proportional (P, current errors), Integral (I, trial history), and Derivative (D-pre, response and regression risk) analyses. The selected plan $c^*$ undergoes contract-constrained editing and staged evaluation, followed by a post-evaluation audit (D-post) and an independent adoption judgment to update the incumbent program $\hat{h}$ and enrich the longitudinal research history.}
\label{pidrsi:fig:overview}
\end{figure*}

The initial program $h_0$ coordinates a report subagent, $K$ sequential CT subagents, and an aggregation and reasoning subagent, corresponding to the upper part of Fig.~\ref{pidrsi:fig:overview}. The report and CT branches produce evidence that the aggregation subagent uses for prediction
\begin{equation}
\begin{aligned}
z_{\mathrm{rep}} &= \mathcal{A}_{\mathrm{rep}}(R),\\
z_{\mathrm{CT}} &= \mathcal{A}_{\mathrm{CT}}^{(1:K)}(V),\\
\hat{\mathbf{y}} &= \mathcal{A}_{\mathrm{agg}}(z_{\mathrm{rep}},z_{\mathrm{CT}}),
\end{aligned}
\label{pidrsi:eq:cliagent}
\end{equation}
where $z_{\mathrm{rep}}$ and $z_{\mathrm{CT}}$ are the outputs of the report and CT branches, respectively. The operator $\mathcal{A}_{\mathrm{CT}}^{(1:K)}$ summarizes the complete CT branch, whose subagents execute in the order shown in the figure. Search revises the program governing evidence interpretation and reasoning while keeping model parameters $\theta$ fixed.

A fixed protocol $\Pi$ specifies the tasks, available evidence, data partitions, model identities, evaluation metrics, and adoption criteria. We instantiate the framework on gastric-cancer prediction. Each search round targets one task $\tau$, evaluated by AUC for pT or pN prediction and full-follow-up C-index for survival prediction. At round $t$, $h_t$ is the parent program to be edited, $\hat{h}_t$ is the currently retained program, and $\PIDRSIEvidence_t$ contains task-evaluation evidence and prior trial records. Initialized with $\hat{h}_0=h_0$, the search updates the retained program through
\begin{equation}
\hat{h}_{t+1}=\mathcal{U}(h_t,\hat{h}_t;\Pi,\PIDRSIEvidence_t),
\label{pidrsi:eq:inner}
\end{equation}
where $\mathcal{U}$ comprises candidate generation, PID analysis, revision, evaluation, and adoption. The notation distinguishes the program used as the editing base from the program used as the adoption reference, although these can be the same program. The next-round arrow in Fig.~\ref{pidrsi:fig:overview} summarizes the continuation of search after the adoption decision.

\subsection{Semantic PID-Guided Candidate Search}

To connect task-level feedback to specific program changes, the Propose module receives the parent $h_t$, current task evidence, and relevant history, and generates a batch $\{c_i\}_{i=1}^{M}$, with $M=5$ by default. Each plan specifies a target problem, a proposed mechanism, the intended edit, expected behavior, and evidence that would support or refute it. The batch is held fixed while the P, I, and D branches analyze it independently and in parallel. We refer to the pre-edit D analysis as D-pre to distinguish it from the post-evaluation D-post analysis.

\paragraph{Current-error analysis (P).}
P examines observed task errors and assesses how each proposed edit addresses a specific performance gap. Its analysis connects the proposed change to the behavior that needs correction.

\paragraph{Historical-evidence analysis (I).}
I draws on the research history in $\PIDRSIEvidence_t$ to examine related trials, recurring failures, previous interventions, and counterexamples. It identifies reusable findings and differences between the current proposal and earlier attempts, drawing on both successful and unsuccessful trials.

\paragraph{Response and risk analysis (D-pre).}
D-pre examines the expected response to an edit, its scope and dependencies, and possible regressions. This analysis identifies behaviors that require attention when evaluating the proposed change.

Each role returns an ordinal score $u_r(c)\in\{0,1,2,3,4\}$ for each plan $c$, together with a rationale, supporting and contradicting evidence, uncertainty, and the expected direction of change. The scores provide an aggregate ranking aid
\begin{equation}
q(c)=\sum_{r\in\mathcal{R}_t}\bar{w}_r u_r(c),
\qquad
\bar{w}_r=\frac{w_r}{\sum_{r'\in\mathcal{R}_t}w_{r'}},
\label{pidrsi:eq:semantic_score}
\end{equation}
where $\mathcal{R}_t\subseteq\{P,I,D\text{-pre}\}$ contains the active roles. We use equal weights by default and renormalize them when a role is removed. A reviewer considers $q(c)$ alongside the complete analyses and may change the ordering with an explicit rationale. The default procedure shortlists two plans for implementation and preliminary evaluation, allows one revised proposal batch, and selects one candidate for full development-set evaluation based on the preliminary results.

\subsection{Behavior-Constrained Revision and Adoption}

\paragraph{Contract-guided revision.}
For each selected plan, a behavior contract specifies the intended change, expected observations, behaviors to preserve, and conditions that would refute the underlying hypothesis. The Edit module uses the plan and contract to revise $h_t$, producing a candidate $h'$. Before evaluation, the candidate is checked against the contract to assess whether the implemented change matches the proposal. This gives the subsequent evaluation explicit behaviors to examine beyond the aggregate task score.

\paragraph{Staged evaluation and D-post analysis.}
Preliminary evaluation examines target behavior and potential regressions to guide candidate selection. The selected candidate then generates fresh predictions on the full development set using the fixed executor $E_{\theta}$, yielding the registered task metric $m_{\tau}(h')$. This full evaluation corresponds to the Evaluate module in Fig.~\ref{pidrsi:fig:overview}. Failed cases and invalid outputs are recorded separately from valid predictions. Full development results provide the performance evidence for adoption.

D-post then compares the original hypothesis and contract with the implemented edit and observed response. It records whether the expected behavior occurred, which observations support the proposal, and which regressions or uncertainties remain. The resulting analysis explains how the measured outcome relates to the intended modification.

\paragraph{Independent adoption.}
An independent AdoptionJudge receives the candidate and retained program's evaluation results directly, together with the D-post analysis. It applies the fixed rubric in $\Pi$ to decide whether the candidate replaces the retained program. For decision $a_t\in\{\mathrm{adopt},\mathrm{keep}\}$, the update is
\begin{equation}
\hat{h}_{t+1}=\begin{cases}
h', & a_t=\mathrm{adopt},\\
\hat{h}_t, & a_t=\mathrm{keep}.
\end{cases}
\label{pidrsi:eq:adoption}
\end{equation}
Candidate ranking allocates evaluation effort, D-post interprets the observed response, and the adoption judge determines which program is retained.

\subsection{Experience Reuse}
\label{sec:experience}
Every evaluated trial, whether adopted or rejected, is archived as a structured record containing its hypothesis, program diff, genealogy, evaluation metrics, D-post audit, and adoption verdict. These records explicitly separate empirical measurements from model-generated interpretations. During subsequent rounds, both the proposal generator and the Integral (I) analyzer retrieve relevant records from this longitudinal memory (via the history-feedback path in Fig.~\ref{pidrsi:fig:overview}) to avoid falsified dead-ends, trace multi-round intervention lineages, and identify when shifted workflow conditions justify revisiting an earlier hypothesis.

\section{Experiments}

\begin{table*}[!t]
\centering
\caption{Main comparison on the development and held-out test sets ($\times 100$). Metrics include pT AUC (T3/T4 vs.\ T1/T2), pN AUC (N+ vs.\ N0), full-follow-up C-index, and their mean (Avg.). Pure LLM and fixed CliAgent baselines are evaluated on the test set only, whereas RSI methods are optimized on the development set first. Bold $\Delta$ entries denote gains of PID-RSI over our initial model $H_0$ (+2.92 Avg., +6.27 C-index).}
\vspace{-0.3cm}
\label{pidinner:tab:performance}
\setlength{\tabcolsep}{6pt}
\renewcommand{\arraystretch}{1.05}
\resizebox{0.9\textwidth}{!}{%
\begin{tabular}{lcccccccc}
\toprule
\multirow{2}{*}{Method}
& \multicolumn{4}{c}{Development}
& \multicolumn{4}{c}{Held-out test}\\
\cmidrule(lr){2-5}\cmidrule(lr){6-9}
& \shortstack{pT AUC $\uparrow$}
& pN AUC $\uparrow$
& C-index $\uparrow$
& Avg.\ $\uparrow$
& \shortstack{pT AUC $\uparrow$}
& pN AUC $\uparrow$
& C-index $\uparrow$
& Avg.\ $\uparrow$\\

\midrule
\multicolumn{9}{l}{\textit{\textbf{Small LLM + pure LLM mode}}}\\
\midrule
Qwen3.6-35B-A3B
& - & - & - & -
& 78.76 & 65.79 & 63.07 & 69.21\\
Qwen3.8-27B
& - & - & - & -
& 78.92 & 68.09 & 68.13 & 71.71\\
Qwen3.8-Flash
& - & - & - & -
& 77.85 & 68.72 & 67.72 & 71.43\\
DeepSeek-V4.1-Flash
& - & - & - & -
& 79.42 & 67.94 & 66.40 & 71.25\\

\midrule
\multicolumn{9}{l}{\textit{\textbf{Closed-source strong LLM + pure LLM mode}}}\\
\midrule
Qwen3.8-Max-0902
& - & - & - & -
& 80.05 & 71.25 & 68.17 & 73.16\\
GPT-6-Luna
& - & - & - & -
& 80.16 & 67.50 & 67.68 & 71.78\\
GPT-6.1-Sol
& - & - & - & -
& 81.12 & 68.89 & 70.05 & 73.35\\
GPT-6-Astra
& - & - & - & -
& 81.91 & 68.44 & 70.30 & 73.55\\

\midrule
\multicolumn{9}{l}{\textit{\textbf{LLMs + our clinical agent (CliAgent) mode}}}\\
\midrule
Qwen3.6-35B-A3B + CliAgent
& - & - & - & -
& 75.64 & 66.66 & 65.85 & 69.38\\
Qwen3.8-27B + CliAgent
& - & - & - & -
& 74.87 & 71.39 & 66.39 & 70.88\\
Qwen3.8-Flash + CliAgent
& - & - & - & -
& 76.92 & 70.82 & 69.54 & 72.43\\
DeepSeek-V4.1-Flash + CliAgent
& - & - & - & -
& 77.39 & 68.98 & 65.18 & 70.52\\
Qwen3.8-Max-0902 + CliAgent
& - & - & - & -
& 78.70 & 70.21 & 69.37 & 72.76\\
GPT-6-Luna + CliAgent
& - & - & - & -
& 78.24 & 66.44 & 69.55 & 71.41\\
GPT-6.1-Sol + CliAgent
& - & - & - & -
& 81.28 & 69.32 & 70.66 & 73.75\\
GPT-6-Astra + CliAgent
& - & - & - & -
& 81.01 & 68.83 & 71.29 & 73.71\\

\midrule
\multicolumn{9}{l}{\textit{\textbf{RSI frameworks optimized on the development set.
Base LLM: Qwen3.6-35B-A3B}}}\\
\midrule
Initial CliAgent for OPRO ($H_0$)
& 80.99 & 66.88 & 66.77 & 71.55
& 76.20 & 68.13 & 61.17 & 68.50\\
OPRO
& 82.41 & 71.44 & 68.11 & 73.99
& 78.26 & 64.82 & 64.32 & 69.13\\
Initial CliAgent for Penguin-RSI ($H_0$)
& 80.99 & 66.88 & 66.77 & 71.55
& 76.20 & 68.13 & 61.17 & 68.50\\
Penguin-RSI
& 85.23 & 68.87 & 68.95 & 74.35
& 77.49 & 65.60 & 63.79 & 68.96\\
Initial CliAgent for ACE ($H_0$)
& 80.99 & 66.88 & 66.77 & 71.55
& 76.20 & 68.13 & 61.17 & 68.50\\
ACE
& 83.34 & 69.88 & 66.71 & 73.31
& 77.86 & 66.21 & 59.99 & 68.02\\

\midrule
\textbf{Baseline $H_0$ (Ours, Qwen3.6-35B-A3B)}
& 82.87 & 66.73 & 64.18 & 71.26
& 77.37 & 65.67 & 59.11 & 67.38\\
\textbf{PID-RSI (Ours)}
& 83.19 & 70.24 & 67.26 & 73.56
& 77.27 & 68.26 & 65.38 & 70.30\\
$\Delta$ (PID-RSI $-$ $H_0$)
& \textbf{+0.32} & \textbf{+3.51} & \textbf{+3.08} & \textbf{+2.30}
& -0.10 & \textbf{+2.59} & \textbf{+6.27} & \textbf{+2.92}\\
\bottomrule
\end{tabular}%
}
\end{table*}

\subsection{Experimental Setup}

\paragraph{Tasks and Data.} Gastric-X~\cite{lu2026gastric} is used for pT and pN staging, with a shared development set of 120 patients (with 37, 24, 28, and 31 cases for T1 to T4, and 61 versus 59 cases for N0 and N+). The internal test sets contain 298 patients for pT (with 94, 59, 69, and 76 cases for T1 to T4) and 299 for pN (with 150 N0 and 149 N+ cases). In addition, an external cohort from our collaborating hospital provides prognosis development and test sets of 108 patients each (46 deaths and 62 censored cases per set), along with 50 cases for external out-of-distribution (OOD) evaluation of both staging tasks.

\paragraph{Search Settings.} Both the execution model ($E_\theta$, Qwen3.6-35B-A3B) and the research model (DeepSeek-Flash), along with the search procedure, are held fixed within each comparison. The reference settings generate $M=5$ candidate plans per round, shortlist two plans for preliminary probing (evaluated on a stratified subset of 20 development cases), allow at most one regenerated batch, and select one candidate for full development-set evaluation per round.

\paragraph{Evaluation Protocol.} We compare the initial task program $h_0$ with the final program selected by Semantic PID-RSI. Development evaluations guide candidate search and adoption. After search, the selected program and $h_0$ are evaluated on the same held-out test set, which is kept strictly separate from candidate generation and selection. We report the primary clinical evaluation metric for each task, with all evaluated final programs maintaining 100\% valid prediction coverage.

\subsection{Overall Performance}

\paragraph{Semantic PID yields the largest held-out improvement.}
With Qwen3.6-35B-A3B fixed as the prediction backbone, \PIDRSIMethod{} raises the three-task test mean from 67.38 to 70.30 (Table~\ref{pidinner:tab:performance}). This 2.92-point gain is more than twice the largest gain achieved by a competing optimizer. Measured from each method's own initial program, ACE gains -0.48 points, OPRO gains 0.63, and Penguin-RSI gains 0.46. Our search starts from a lower test mean than the three competing searches (67.38 versus 68.50) and finishes with the highest mean. The improvement comes from the two tasks on which the initial CliAgent is weakest. Held-out pN AUC rises from 65.67 to 68.26, and survival C-index rises from 59.11 to 65.38. 

\paragraph{Which development gains transfer to test patients?}
Nodal staging illustrates the feedback ambiguity that motivates Semantic PID. OPRO, Penguin-RSI, ACE, and \PIDRSIMethod{} all improve pN AUC on development, yet the selected revisions have different effects on held-out patients. OPRO achieves the largest development gain, from 66.88 to 71.44, while its test AUC falls from 68.13 to 64.82. Penguin-RSI gains 1.99 points on development and loses 2.53 on test, while ACE gains 3.00 points on development and loses 1.92 on test. In contrast, \PIDRSIMethod{} gains 3.51 points on development and 2.59 on test, yielding the highest held-out pN AUC among the four optimizers. Survival shows a second advantage: the 6.27-point increase over our initial program exceeds the corresponding changes for ACE (-1.18), OPRO (+3.15), and Penguin-RSI (+2.62). Across these two endpoints, the revisions selected with Semantic PID yield larger held-out improvements over their own initialization than the competing optimizers.

\paragraph{Comparison with direct inference and fixed clinical agents.} On the held-out cohorts, evolving CliAgent with the 3B-activated Qwen3.6-35B-A3B backbone achieves a three-task mean of 70.30. Compared with larger models under direct inference, our program attains 98.67\% and 98.03\% of the mean scores of DeepSeek-V4.1-Flash (71.25) and Qwen3.8-27B (71.71), respectively, while reaching 95.58\% of the strongest direct-inference proprietary model, GPT-6-Astra (73.55). These results indicate that our evolved 3B-activated agent closely approaches the performance of substantially stronger models.

Furthermore, equipping GPT-6.1-Sol with our fixed CliAgent raises its mean score from 73.35 to 73.75, establishing the strongest fixed-agent baseline. Against this upper bound, our evolved 3B-activated program (70.30) still retains 95.32\% of the overall performance. At the task level, Semantic PID-RSI maintains stable tumor-depth staging performance (77.27 versus 77.37 initial pT AUC, reaching 94.34\% of the 81.91 best baseline by GPT-6-Astra), while substantially boosting held-out pN AUC by 2.59 points and survival C-index by 6.27 points. Achieving substantial gains over initialization while retaining over 95\% of the strongest proprietary baselines confirms the effectiveness of Semantic PID-RSI on a lightweight backbone.

\subsection{Out-of-Distribution generalization} To assess the out-of-distribution (OOD) generalization of the framework evolved through PID-RSI, we evaluate it on an external T/N staging dataset from a collaborating hospital. As shown in Table \ref{pidinner:tab:ood}, the evolved framework improves pT and pN AUC by 6.01 and 8.78 points, respectively, over the initial framework. These results support the effectiveness of PID-RSI under distribution shift.

\begin{figure}
    \centering
    \includegraphics[width=\linewidth]{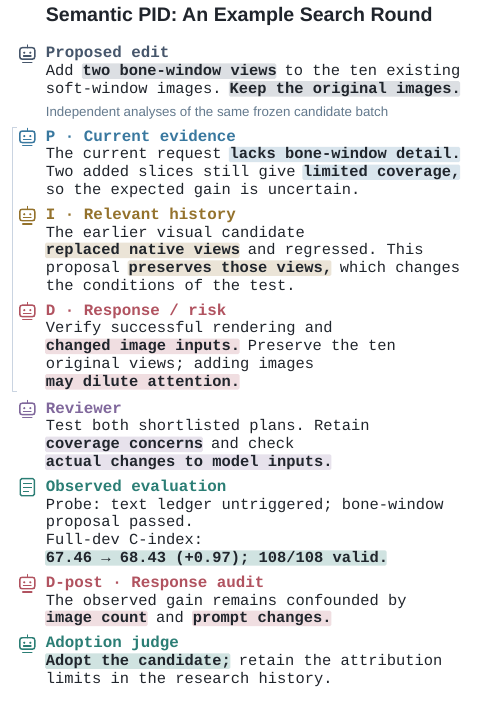}
    \vspace{-0.3cm}
    \caption{\textbf{An adopted revision in Semantic PID-RSI.} Adding two bone-window views while preserving existing images improves the development-set C-index by 0.97 points. The judge adopts the revision while retaining the attribution limits identified by D-post.}
    \label{fig:single_column}
\end{figure}

\subsection{Case study of an adopted revision}
In round R03, Semantic PID-RSI evaluated a proposal to add two bone-window views while preserving the ten existing soft-tissue-window images, as shown in Fig. \ref{fig:single_column}. The three roles independently assessed the same candidate batch. \textbf{P} identified missing bone-window detail but noted that the limited coverage of two additional slices made the expected benefit uncertain. \textbf{I} distinguished this proposal from an earlier revision that replaced the original views and degraded performance. \textbf{D-pre} highlighted possible attention dilution and requested verification of image rendering and actual model inputs. Following reviewer selection, the proposal passed preliminary evaluation and improved the full development-set C-index from 67.46 to 68.43 (+0.97 points), with valid outputs for all 108 cases. \textbf{D-post} nevertheless flagged concurrent changes in image count and prompting, which prevented attributing the gain solely to bone-window information. The independent adoption judge accepted the candidate while preserving these attribution limits in the research history.

\subsection{Ablation Studies}

\paragraph{Current errors and trial history provide complementary feedback.}
Table~\ref{pidinner:tab:ablation} tests which feedback sources account for the improvement in survival prediction. Each search starts from the same CliAgent and uses the same prediction model, patient splits, and evaluation budget. The full method raises held-out C-index from 59.11 to 65.38. Removing historical-evidence analysis (I) reduces the score to 60.15, the largest drop of any ablation (5.23 points). With access to current errors and response analysis, this variant gains only 1.04 points over initialization. Removing current-error analysis (P) produces a similarly large drop of 4.79 points, where the remaining I and D roles reach 60.59. These results support the two distinct questions built into the improver. P identifies which failures in the current program merit intervention, while I provides evidence about what happened when related changes were tried. Omitting either source leaves most of the full method's prognosis gain unrealized.

\paragraph{Response analysis improves on the strongest two-role configuration.}
The strongest two-role search retains P and I. It reaches 63.24 C-index, 4.13 points above the initial program. Restoring D raises the score by a further 2.14 points to 65.38. This comparison addresses the second challenge in the introduction: even after the improver identifies a recurring error and reviews earlier attempts, it must still judge how a proposed edit will affect the program. D-pre examines the expected reach of the edit and the behaviors at risk of regression. D-post compares those expectations with the observed response before independent adoption. The ablation removes both stages together. Their inclusion yields the best held-out survival ranking across all four configurations. Overall, on this challenging long-term prognosis endpoint, ablating historical-evidence analysis (I) induces an even steeper degradation of 5.23 points compared with 4.79 points when removing current-error analysis (P), which underscores the critical role of longitudinal trial memory in noisy clinical optimization.

\begin{table}[t]
\centering
\caption{
External evaluation on the OOD cohort ($n=50$).
}
\label{pidinner:tab:ood}
\vspace{-0.3cm}
\resizebox{0.5\columnwidth}{!}{%
\begin{tabular}{lcc}
\toprule
Method & pT $\uparrow$ & pN $\uparrow$ \\
\midrule
$H_0$     & 69.18          & 57.39          \\
PID-RSI & \textbf{75.19} & \textbf{66.17} \\
\midrule
$\Delta$   & +6.01          & +8.78          \\
\bottomrule
\end{tabular}%
}
\end{table}

\begin{table}[!t]
\centering
\caption{Ablation of Semantic PID roles on prognosis prediction.
Checkmarks indicate enabled roles, and $H_0$ denotes the initial model
before optimization. C-index is evaluated over the full follow-up.}
\vspace{-0.3cm}
\label{pidinner:tab:ablation}
\resizebox{\columnwidth}{!}{%
\begin{tabular}{lccccc}
\toprule
Role & $H_0$ & w/o P & w/o I & w/o D & Full PID-RSI\\
\midrule
P & $\times$ & $\times$ & $\checkmark$ & $\checkmark$ & $\checkmark$\\
I & $\times$ & $\checkmark$ & $\times$ & $\checkmark$ & $\checkmark$\\
D & $\times$ & $\checkmark$ & $\checkmark$ & $\times$ & $\checkmark$\\
\midrule
C-index $\uparrow$ & 59.11 & 60.59 & 60.15 & 63.24 & \textbf{65.38}\\
\bottomrule
\end{tabular}}
\end{table}

\begin{figure}[t]
    \centering
    \includegraphics[width=\columnwidth]{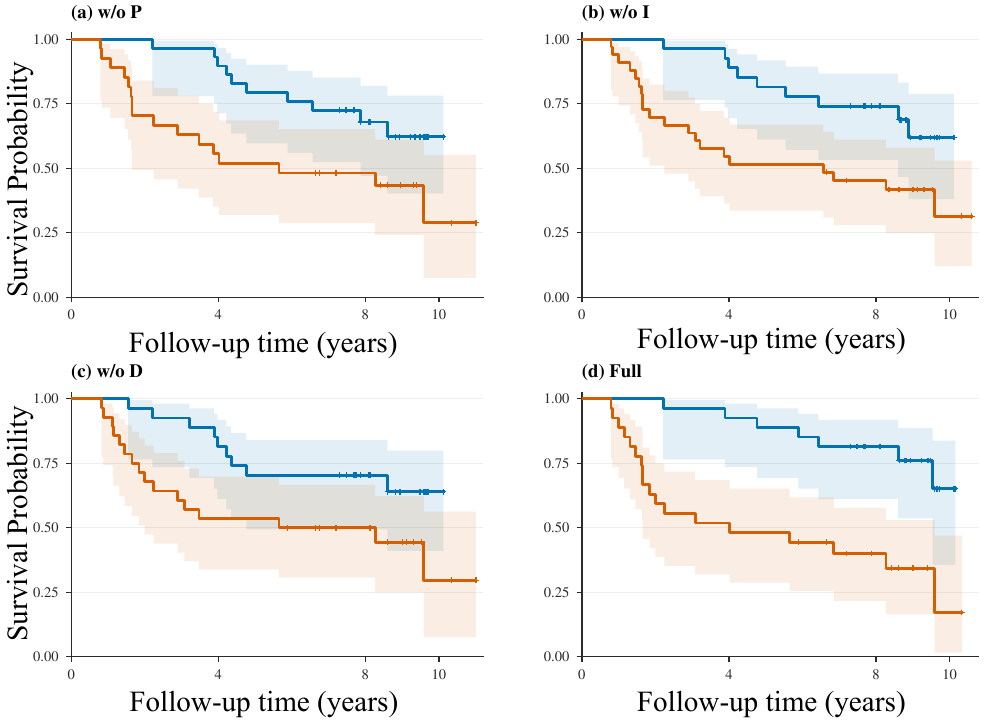}
    \vspace{-0.3cm}
    \caption{
        \textbf{Kaplan--Meier curves for the prognosis ablation study.}
        Results are shown for (a) w/o P, (b) w/o I, (c) w/o D,
        and (d) the full model.
        Blue and orange curves represent the lowest and highest
        predicted-risk quartiles, respectively, with cutoffs computed
        separately for each setting on the common held-out cohort
        ($n=108$).
        Intermediate-risk patients are omitted from these plots,
        and tied scores are kept together.
        Shaded regions indicate 95\% pointwise confidence intervals,
        and crosses mark censored observations.
        Risk grouping is post hoc and exploratory.
    }
    \label{fig:prognosis_ablation_km}
\end{figure}

\subsection{Mechanism Analysis}

\begin{figure}[t]
    \centering
    \includegraphics[width=\columnwidth]{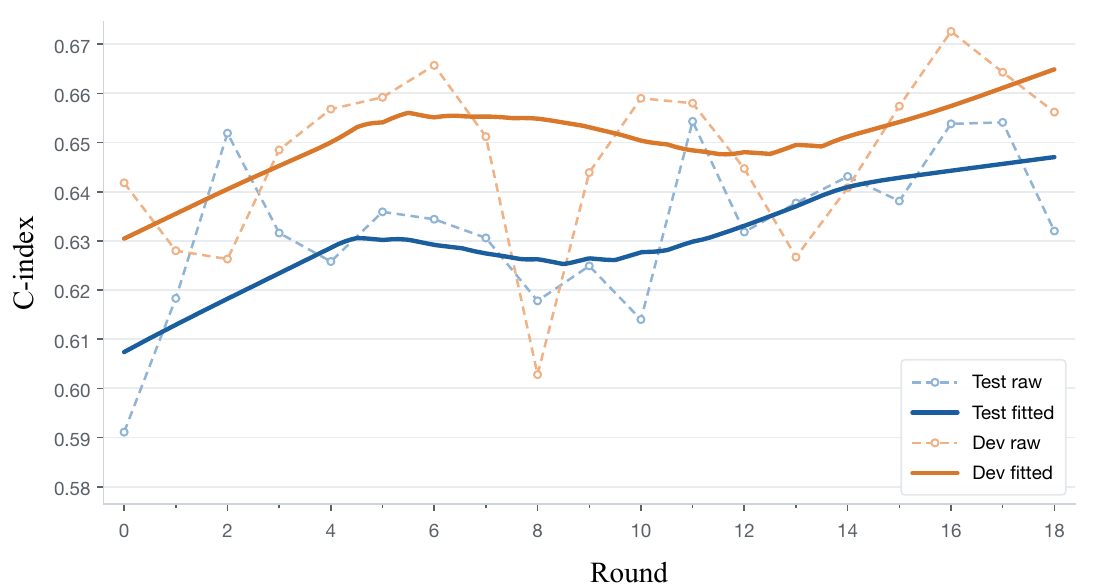}
    \vspace{-0.3cm}
    \caption{
        \textbf{Survival optimization trajectory across search rounds.}
        Orange and blue curves denote development and held-out test full-follow-up C-index, respectively, from initialization ($H_0$, Round~0) to Round~18.
        Dashed lines with hollow markers show raw candidate scores at each round (`raw'), while bold solid curves depict smoothed progression trends (`fitted').
        Despite exploratory fluctuations (e.g., Round~8), both development and test trends rise consistently without divergence, reaching the final adopted program at Round~16 (Dev: 0.6726, Test: 0.6538 vs.\ 0.5911 at Round~0).
    }
    \label{fig:pid_traj}
\end{figure}

\paragraph{Search Trajectories.}
Figure~\ref{fig:pid_traj} shows the survival candidates at initialization and even-numbered search rounds. The two curves report development and held-out full-follow-up C-index for the same candidates. Lines connect the displayed observations, and stars identify the final selected candidate from round 16. This view contrasts the evaluated alternatives with the program retained at the end of search.

The displayed development scores decrease to 0.6028 at round 8 and reach 0.6726 at round 16. The final selected program has a held-out C-index of 0.6538, compared with 0.5911 for $H_0$. At round 18, the candidate scores are 0.6562 on development and 0.6320 on the test set, while the round-16 program remains the final incumbent. These observations distinguish exploration of new candidates from replacement of the retained program.

\FloatBarrier

\section{Conclusion}

We presented a unified agentic architecture and self-improvement framework for high-uncertainty vertical medical decision-making. We first orchestrated {CliAgent} to ground clinical reasoning in structured patient evidence, and introduced {Semantic PID-RSI} to recursively evolve medical agents by {improving the improver}. By evaluating candidate modification batches through complementary current-error ({P}), historical-evidence ({I}), and response-risk ({D-pre}/{D-post}) analyses, and enforcing disciplined updates via {behavior contracts}, staged probing, and independent adoption judging, our approach resolves feedback ambiguity while resisting development-set overfitting. Instantiated on postoperative pathological staging and survival prognosis, evolving {CliAgent} with {Semantic PID-RSI} empowers a lightweight {3B-activated} open-weight model to outperform existing RSI baselines and approach frontier proprietary models equipped with {CliAgent}.

\clearpage
\section*{LLM Usage Statement}
The authors take full responsibility for all submitted content, and no full manuscript or major parts were provided to an LLM for reviewer-style feedback. Generative AI tools were used in this work at two levels of involvement: at the Assist level (editing or refining author-produced content) for writing and text editing, related work and citations, experiments including design, execution, and analysis, figures and visual material, and research ideas, hypotheses, claims, or conclusions; and at the Generate level (where authors reviewed, edited, and approved tool-generated output through one-shot or interactive use) for code and implementation, as well as data preparation including synthetic data, dataset creation, and annotation.
{
    \small

    \IfFileExists{ieeenat_fullname.bst}{
        \bibliographystyle{ieeenat_fullname}
    }{
        \bibliographystyle{ieee_fullname}
    }

    \bibliography{sec/references}
}

\end{document}